\documentclass[conference, letterpaper]{IEEEtran}

\ifCLASSINFOpdf
\else
\fi
\ifCLASSINFOpdf
   \usepackage[pdftex]{graphicx}
\else
\fi

\usepackage[cmex10]{amsmath}
\usepackage{color}
\usepackage{fancyhdr}
\usepackage[caption=false,font=footnotesize]{subfig}

\usepackage{tikz}
\usepackage{commath}
\usetikzlibrary{positioning, fit, arrows.meta, shapes, arrows}
\usetikzlibrary{calc}

\renewcommand{\thispagestyle}[2]{}

\fancypagestyle{plain}{
        \fancyhead{}
        \fancyhead[C]{first page center header}
        \fancyfoot{}
        \fancyfoot[C]{first page center footer}
}
\begin{document}

%
\title{Comparative evaluation of CNN architectures for Image Caption Generation}

\author{\IEEEauthorblockN{Sulabh Katiyar}
\IEEEauthorblockA{Department of\\ Computer Science and Engineering\\
National Institute of Technology, Silchar\\
Assam, India 788010\\
}
\and
\IEEEauthorblockN{Samir Kumar Borgohain}
\IEEEauthorblockA{Department of\\ Computer Science and Engineering\\
National Institute of Technology, Silchar\\
Assam, India 788010\\
}}

\maketitle

\begin{abstract}
Aided by recent advances in Deep Learning, Image Caption Generation has seen tremendous progress over the last few years. Most methods use transfer learning to extract visual information, in the form of image features, with the help of pre-trained Convolutional Neural Network models followed by transformation of the visual information using a Caption Generator module to generate the output sentences. Different methods have used different Convolutional Neural Network Architectures and, to the best of our knowledge, there is no systematic study which compares the relative efficacy of different Convolutional Neural Network architectures for extracting the visual information. In this work, we have evaluated 17 different Convolutional Neural Networks on two popular Image Caption Generation frameworks: the first based on Neural Image Caption (NIC) generation model and the second based on Soft-Attention framework. We observe that model complexity of Convolutional Neural Network, as measured by number of parameters, and the accuracy of the model on Object Recognition task does not necessarily co-relate with its efficacy on feature extraction for Image Caption Generation task. We release the code at https://github.com/iamsulabh/cnn\_variants
\end{abstract}


\IEEEpeerreviewmaketitle

\section{Introduction}
\label{introduction}
Image Caption Generation involves training a Machine Learning model to learn to automatically produce a single sentence description for an image. For human beings it is a trivial task. However for a Machine Learning method to be able to perform this task, it has to learn to extract all the relevant information contained in the image and then to convert this visual information into a suitable representation of the image which can be used to generate a natural language sentence description of the image. The visual features extracted from the image should contain information about all the relevant objects present in the image, the relationships among the objects and the activity settings of the scene. Then the information needs to be suitably encoded, generally in a vectorized form, so that the sentence generator module can convert this into a human readable sentence. Furthermore, some information may be implicit in the scene such as a scene where a group of football players are running in a football field but the football is not present in the scene frame. Thus the model may need to learn some level of knowledge about the world as well. However, the ability to automate the caption generation process has many benefits for the society as it can either replace or complement any method that seeks to extract some information from the images and has applications in the fields of education, military, medicine, etc., as well as applications in some specific problems such as helping visually impaired people in navigation or generating news information from images.

During the last few years there has been tremendous progress in Image caption generation due to advances in Computer Vision and Natural Language Processing domains. The progress made in Object Recognition task due to availability of large annotated datasets such as ImageNet \cite{imageneta} has led to availability of pre-trained Convolutional Neural Network (CNN) models which can extract useful information from the image in vectorized form which can then be used by caption generation module (called the decoder) to generate caption sentences. Similarly, progress in solving machine translation with methods such as encoder-decoder framework proposed in \cite{sutskever}, \cite{cho} has led to adoption of similar format for Image Caption Generation where the source sentence in machine translation task is replaced by the image in caption generation task and then the process is approached as 'translation' of image to sentence, as has been done in works such as \cite{karpathy},\cite{mao},\cite{vinyals}. The attention based framework proposed by \cite{bahdanau} where the decoder learns to focus on certain parts of the source sentence at certain time-steps has been adapted in caption generation in such as way that the decoder focuses on portions of image at certain time-steps \cite{attendtell}. A detailed survey of Image Caption Generation has been provided in \cite{bernardi} and \cite{hossain}. 

Although there has been a lot of focus on the decoder which 'interprets' the image features and 'translates' them into a caption, there has not been enough focus on the encoder which 'encodes' the source image into a suitable visual representation (called image features). This is mainly because most methods use transfer learning to extract image features from pre-trained Convolutional Neural Networks (CNN) \cite{lecun} which are trained on the Object Detection task of the ImageNet Large Scale Visual Recognition Challenge \cite{imagenetvisualtask} where the goal is to predict the object category out of 1000 categories annotated in the dataset. Since the last layer of the CNN produces a 1000 length vector containing relative probabilities of all object categories, the last layer is dropped and the output(s) of intermediate layer(s) is(are) used as image features. Numerous CNN architectures have been proposed with varying complexity and efficacy and many have been utilized for Image Caption Generation as well. However, this makes it difficult to undertake a fair comparison of Image Caption Generation methods since the difference in performance could be either due to difference in effectiveness of decoders in sentence generation or due to difference in effectiveness of encoders in feature extraction. 

Hence, in this work we evaluate Image Caption Generation using popular CNN architectures which have been used for Object Recognition task and analyse the co-relation between model complexity, as measured by the total number of parameters, and the effectiveness of different CNN architectures on feature extraction for Image Caption Generation. We use two popular Image Caption Generation frameworks: (a) Neural Image Caption (NIC) Generator proposed in \cite{vinyals} and (b) Soft Attention based Image Caption Generation proposed in \cite{attendtell}. 
We observe that the performance of Image Caption Generation varies with the use of different CNN architectures and is not directly correlated with either the model complexity or performance of CNN on object recognition task. To further validate our findings, we evaluate multiple versions of ResNet CNN \cite{resnet} with different depths (number of layers in the CNN) and complexity: ResNet18, ResNet34, ResNet50, ResNet101, ResNet152 where the numerical part in the name stands for the number of layers in the CNN (such as 18 layers in ResNet18 and so on). We evaluate multiple versions of VGG CNN \cite{vgg16} architecture: VGG-11, VGG-13, VGG-16 and VGG-19 and multiple versions of DenseNet CNN \cite{densenet} architecture: Densenet121, Densenet169, DenseNet201 and Densenet161, each of which has different number of parameters. We observe that performance does not improve with the increase in the number of layers, and consequently, increase in model complexity. This further validates our observation that effectiveness of CNN architectures for Image Caption Generation depends on the model design and that the model complexity or the performance on Object Detection task are not good indicators of effectiveness of CNN for Image Caption Generation.
To the best of our knowledge, this is the first such detailed analysis of the role of CNN architectures as image feature extractors for Image Caption Generation task. In addition, to further the future research work in this area, we also make the implementation code\footnote{https://github.com/iamsulabh/cnn\_variants} available for reference.

This paper is divided into following sections: In Section \ref{related_word},we discuss the relevant methods proposed in the literature, in Section \ref{proposed_method}, we discuss the methodology of our work, in Section \ref{results} we present and discuss the experimental results and in Section \ref{conclusion} we discuss the implications of our work and possible future studies.

\section{Related Work}
\label{related_word}
Some of the earliest works attempted to solve the problem of caption generation in constrained environments such as the work proposed in \cite{kojima} where the authors try to generate captions for objects present in an office setting. Such methods had limited scalability and applications. Some works tried to address the task as a \textit{Retrieval problem} where a pool of sentences was constructed which could describe all (or most) images in a particular setting. Then for a target image, a sentence which was deemed appropriate by the algorithm was selected as the caption. For example, in \cite{farhadi}, the authors construct a 'meaning space' which consists of triplets of $<$objects, actions, scene$>$. This is used as a common mapping space for images and sentences. A similarity measure is used to find sentences with the highest similarity to the target image and the most similar sentence is selected as the caption. In \cite{mason}, a set of images are retrieved from the training data which are similar to the target image using a visual similarity measure. Then a word probability density conditioned on the target image is calculated using the captions of the images that were retrieved in the last step. Then the captions in the dataset are scored using this word probability density and the sentence which has the highest score is selected as the caption for the target image. The retrieval based methods generally produce grammatically correct and fluent captions because they select human generated sentence for a target image. However, this approach is not scalable because a large number of sentences need to be included in the pool for each kind of environment. Also the selected sentence may not even be relevant because the same kind of objects may have different kind of relationships among them which cannot be described by a fixed set of sentences.

Another class of approaches are the \textit{Template based methods} which construct a set of hand-coded sentence templates according to the rules of grammar and semantics and optimization algorithms. Then the methods plug in different object components and their relationships into the templates to generate sentences for the target image. For example, in \cite{kulkarni}, Conditional Random Fields are used to recognize image contents. A graph is constructed with the image objects, their relationships and attributes as nodes of the graph. The reference captions available with the training images are used to calculate pairwise relationship functions using statistical inference and the visual concepts are used to determine the unary operators on the nodes. In \cite{li}, visual models are used to extract information about objects, attributes and spatial relationships. The visual information is encoded in the form of [$<$adjective1,object1$>$,preposition,$<$adjective2,object2$>$] triplets. Then n-gram frequency counts are extracted from web-scale training dataset using statistical inference. Dynamic programming is used to determine optimal combination of phrases to perform phrase fusion to construct the sentences. Although the Template based approaches are able to generate more varied captions, they are still handicapped by the problems of scalability because a large number of sentence templates are to be hand-coded and even then a lot of phrase combinations may be left out.

In recent years, most of the works proposed in the literature have employed Deep Learning to generate captions. Most works use CNNs, which are pre-trained on the ImageNet Object Recognition dataset \cite{imageneta}, to extract vectorized representation of the image. Words of a sentence are represented as Word Embedding vectors extracted from a look-up table. The look up table is learned during training as the set of weights of the Embedding Layer. The image and word information is combined in different ways. Most methods use different variants of Recurrent Neural Network \cite{rnn} (RNN) to model the temporal relationships between words in the sentence. In \cite{mao}, the image features extracted from CNN and the word embeddings are mapped to the same vector space and merged using element-wise addition at each time-step. Then the merged image features and word embeddings are used as input to a MultiModal Reccurent Neural Network (m-RNN) which generates the output. The authors use AlexNet\cite{alexnet} and VGG-16 \cite{vgg16} as CNNs to extract image features. In \cite{karpathy} a Bidirectional Recurrent Neural Network is used as decoder because it can map the word relationships with both the words that precede and the words that succeed a particular word in the sentence. The word embeddings and image features are merged before being fed into the decoder. The authors use AlexNet \cite{alexnet} CNN to extract image features. In \cite{vinyals}, a Long Short Term Memory Network \cite{lstm} is used as decoder. The image features are mapped to the vector space spanned by hidden state representations of the LSTM and are used as initial hidden state of the LSTM. Thus the image information is fed to LSTM at initial state only. The LSTM takes in previously generated words as input (with a special 'start token' as the first input) and generates the next word sequentially. The authors use \cite{inception} as CNN for extracting image features. Using the Attention approach, in \cite{attendtell} the authors train the model to focus on certain parts of the image at certain time-steps. This attention mechanism takes as input, the image features and output until the last time-step and generates an image representation conditioned on text input. This is merged with the word embeddings at the current time-step by using vector concatenation operation and used as input to the LSTM generator. The authors used VGGNet \cite{vgg16} CNN as image feature extractor. Recently, methods using Convolutional Neural Networks as sequence generators have been proposed such as in \cite{gehring} for text generation. Based on this approach, \cite{aneja} propose a method which uses a CNN for encoding the image and another CNN for decoding the image. The CNN decoder is similar to the one used in \cite{gehring} and uses a hierarchy of layers to model word relationships of increasing complexity. The authors use ResNet152\cite{resnet} CNN to encode the image features. More recently, Transformer Network has been used which uses self-attention to model word relationships instead of Recurrent or Convolutional operations \cite{vaswani}. Based on this approach a Transformer based caption generation is proposed in \cite{capstransformer}. Since most of the methods use different CNN architectures to extract image features, there is a need for a comparative analysis of their effectiveness in image feature extraction using the same overall format for caption generation.

\section{Proposed Method}
\label{proposed_method}
In image caption generation, given an image the task is to generate a set of words $S = \{w_1, w_2, w_3, ..., w_{L}\}$ where $w_i \in \mathcal{V}$ where $L$ is the length of the sentence and $\mathcal{V}$ represents the vocabulary of the dataset. The words $w_1$ and $w_{L}$ are usually the special tokens for start and end of the sentence. Two more special tokens for 'unknown' and 'padding' are also used for representing unknown words (which may be the stop words and rare words that have been removed from dataset to speed up training) and padding the end of the sentence (to make all sentences of equal length because RNNs do not handle sentence of different lengths in the same batch), respectively. Given pairs of image and sentence, ($I_N, S_i$) for $i \in (1,2,3,...,j)$, during training we maximize the probability $P(S_i | I_N, \theta)$ where $j$ is the number of captions for an image in training set and $\theta$ represents the set of parameters in the model. Hence, as mentioned in \cite{vinyals}, during training the model learns to update the set of parameters $\theta$ such that the probability of generation of correct captions is maximized according to the equation,
\begin{equation}\label{eq1}
    \theta^\star = arg max \sum_{(I,S)} log_p(S|I,\theta)
\end{equation}
where $\theta$ is the set of all parameters of the model, $I$ is the image and $S$ is one of the reference captions provided with the image.
We can use chain rule because generation of words of a sentence depends on previously generated words, and hence Equation \ref{eq1} can be extended to the constituent words of the sentence as,
\begin{equation}\label{eq2}
    log_p(S|I, \theta) = \sum_{t=0}^{L}log_p(w_t|I, \theta, w_1, w_2, ..., w_{t-1})
\end{equation}
where \begin{math} w_1, w_2, ..., w_{L} \end{math} are the words in the sentence '$S$' of length $L$. This equation can be modelled using a Recurrent Neural Network which generates the next output conditioned on the previous words of the sentence. We have used LSTM as the RNN variant for our experiments.

\begin{figure*}[h]
\tikzstyle{startstop} = [rectangle, rounded corners, minimum width=1cm, minimum height=1cm,text centered, draw=black]
\tikzstyle{io} = [trapezium, trapezium left angle=70, trapezium right angle=110, minimum width=1cm, minimum height=.5cm, text centered, draw=black]
\tikzstyle{input} = [rectangle, minimum width=0.5cm, minimum height=0.5cm, text centered, draw=black]
\tikzstyle{LSTM} = [rectangle, rounded corners, minimum width=1cm, minimum height=1.0cm, text centered, draw=black, fill=blue!20]
\tikzstyle{SOFT} = [rectangle, rounded corners, minimum width=2cm, minimum height=1cm, text centered, draw=black, fill=blue!10]
\tikzstyle{decision} = [diamond, minimum width=1cm, minimum height=.2cm, text centered, draw=black]
\tikzstyle{fcimagestyle} = [rectangle, rounded corners, minimum width=1cm, minimum height=1cm,text centered, draw=black, fill=red!30]
\tikzstyle{fctextstyle} = [rectangle, rounded corners, minimum width=1cm, minimum height=1cm,text centered, draw=black, fill=orange!20]
\tikzstyle{embedstyle} = [rectangle, rounded corners, minimum width=0.5cm, minimum height=0.5cm,text centered, draw=black, fill=blue!10]
\tikzstyle{NASNET} = [circle, rounded corners, minimum width=0.8cm, minimum height=0.8cm,text centered, draw=black, fill=gray!30]
\tikzstyle{dotbox} = [rectangle, rounded corners, minimum width=1cm, minimum height=1cm,text centered, draw=none]
\tikzstyle{atimagestyle} = [circle, rounded corners, minimum width=0.7cm, minimum height=0.7cm,text centered, draw=black, fill=olive!20]

\tikzstyle{arrow} = [thick,->]
\centering
\begin{tikzpicture}[node distance = 1.0cm]
    \node (lstm1) [LSTM]{\small LSTM};
     \node (soft) [input,above of=lstm1,node distance=1.0cm] {\small SOFTMAX};
    \node (output) [input,above of=soft,node distance=0.75cm] {\small $w_0,w_1,..., w_n$};
    \node (cnn)[NASNET, left of=lstm1, node distance=1.7cm, yshift=-1.25cm] {\small CNN};
    \node (image) [input,below of=cnn,node distance=1.0cm] {\small image};
    \node (emb) [embedstyle,below of=lstm1,node distance=1.25cm] {\small Embedding};
    \node (words) [input,below of=emb,node distance=1.0cm] {\small text input};
    \node (dota) [dotbox, below of=cnn, node distance=2.1cm] {(a)};
    \node (lstm2) [LSTM, right of = lstm1, node distance = 10cm]{\small LSTM};
    \node (soft2) [input,above of=lstm2,node distance=1.0cm, xshift = 0.5cm] {\small SOFTMAX};
    \node (output2) [input,above of=soft2,node distance=0.75cm, xshift = 0.0cm] {\small $w_0,w_1,..., w_n$};
    \node (emb2) [embedstyle,below of=lstm2,node distance=1.25cm] {\small Embedding};
    \node (cnn2)[NASNET, left of=lstm2, node distance=2.25cm, yshift=-1.5cm] {\small CNN};
    \node (image2) [input,below of=cnn2,node distance=1.0cm] {\small image};
    \node (words2) [input,below of=lstm2,node distance=2.5cm] {\small text input};
    \node (dotb) [dotbox, below of=cnn2, node distance=2.0cm] {(b)};
    \node (att)[atimagestyle, above of=cnn2, node distance=2.0cm,xshift=0.5cm] {\small AL};

    \draw [arrow, ->] ([xshift = -0.2cm]cnn.north) |- (lstm1.west);
    \draw [arrow, ->] (words.north) |- (emb.south);
    \draw [arrow, ->] (emb.north) |- (lstm1.south);
    \draw [arrow, ->] (lstm1.north) |- (soft.south);
    \draw [arrow, ->] (soft.north) |- (output.south);
    \draw [arrow, ->] (image.north) |- (cnn.south);
    \draw [arrow, ->] (lstm1) to [out=30,in=340,looseness=4] (lstm1);

    \draw [arrow, ->] ([xshift = -0.2cm]cnn2.north) |- (att.west);
    
    \draw [arrow, ->] ([xshift = 0.2cm]cnn2.north) |- ([yshift=-0.3cm]lstm2.west);
    \draw [arrow, ->] ([xshift = 0.2cm]lstm2.north) -- ([xshift = -0.3cm]soft2.south);
    \draw [arrow, ->] (words2.north) -- (emb2.south);
    \draw [arrow, ->] (emb2.north) -- (lstm2.south);
    \draw [arrow, ->] (soft2.north) -- (output2.south);
    \draw [arrow, ->] (image2.north) -- (cnn2.south);
    \draw [arrow, ->] (lstm2) to [out=30,in=340,looseness=4] (lstm2);
    \draw [arrow, ->] (att) to [out=-330,in=-230,looseness=1] (lstm2);
    \draw [arrow, ->] (lstm2) to [out=180,in=-40,looseness=1] (att);

\end{tikzpicture}
\centering
\caption{An overview of the two approaches proposed in this work: \\(a) Encoder-Decoder based approach. 
(b) Attention based approach with an attention mechanism to focus on salient portions of the image.\\
(AL stands for Attention Layer)}
\label{overview}
\end{figure*}
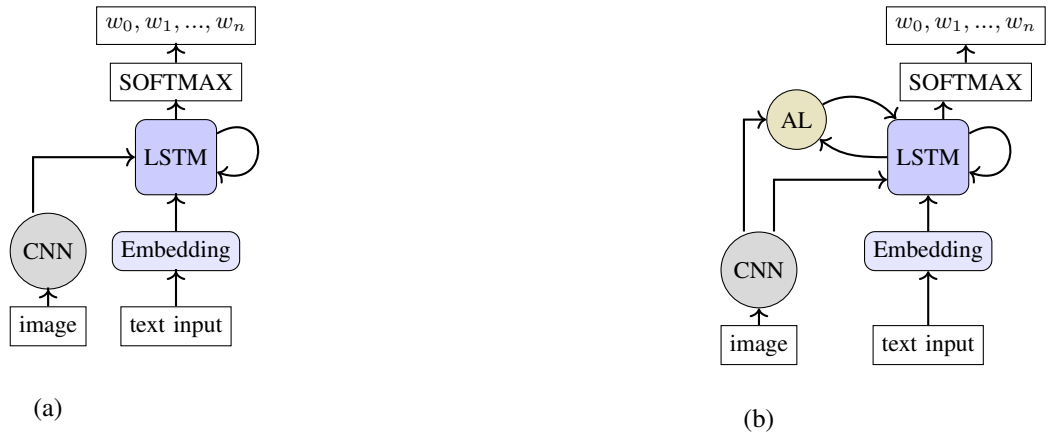

In this work, we evaluate caption generation performance on two popular encoder-decoder frameworks with certain modifications. For both the methods, we experiment with different CNN architectures for image feature extraction and analyse the effects on performance.\\
The first method is based on Neural Image Caption Generation method proposed in \cite{vinyals}. However, unlike the method proposed in \cite{vinyals}, we have not used model ensembles to improve performance. In addition, we have extracted image features from a lower layer of the CNN which generates a set of vectors each of which contain information about a region of the image. We have observed that this leads to better performance as the decoder is able to use region specific information to generate captions. Throughout this paper, this will be referred to as 'CNN+LSTM' approach with the word 'CNN' replaced by the name of CNN architecture used in the experiment. For example, 'ResNet18+LSTM' refers to caption generation with ResNet18 as the CNN.\\
The second method is similar to the Soft Attention method proposed in \cite{attendtell}. We use an attention mechanism which learns to focus on certain portions of image for at certain time-steps for generating the captions. Similar to the CNN+LSTM approach, this Soft Attention approach will be referred as 'CNN+LSTM+Attention' approach with the word 'CNN' replaced by the name of CNN architecture used. Figure \ref{overview} explains both the methods.

\subsection{Image Feature extraction}
\label{image_feature_extraction}
For extracting image features, we use CNNs which were pre-trained on ImageNet datset \cite{imageneta} for the Imagenet  Large  Scale  Visual Recognition Challenge \cite{imagenetvisualtask}. The models generate a single output vector containing the relative probabilities of different object categories (with 1000 categories in total). We remove this last layer from the CNN since we need more fine-grained information. Also, we remove all the layers at the top (with the input layer being called the bottom layer) which produce a single vector as output because we need a set of vectors as output which contain information about different regions of the image.
Hence, the image features are a set of vectors denoted as, \begin{math} \mathbf{a} = \{ a_1,a_2,a_3,...a_{\vert a \vert} \}, a_i \in \mathcal{R}^{D} \end{math} where $\vert a \vert$ is the number of feature vectors contained in $\mathbf{a}$, $\mathcal{R}$ represents real numbers and $D$ is dimension of each vector. For example, ResNet152 CNN \cite{resnet} generates a set of 8, 2048 dimensional vectors.

The set of image feature vectors thus generated are used in two ways in the methods used in this work. In the 'CNN+LSTM' method, the image features are mapped to the vector space of hidden state of the LSTM and used to initialize the hidden and cell state of the LSTM decoder. For the 'CNN+LSTM+Attention' method, in addition to hidden and cell state initialization, the set of image feature vectors is also used at each time-step to calculate attention weighted image features which contain information from those regions in the image which are important at the current time-step. We explain this in detail in Sections \ref{cnn_lstm} and \ref{cnn_lstm_attention}.

\subsection{CNN + LSTM method}
\label{cnn_lstm}
In this method, we use a CNN encoder to extract image information and use that information as the initial hidden state of the LSTM decoder. Using the set of image feature vectors thus obtained as described in Section \ref{image_feature_extraction}, we obtain a single vector by averaging the values of all vectors in the set as,
\begin{equation}
a_{ave} = \sum_i^{\vert \mathbf{a} \vert} a_i, i\in (1,2,...,\vert \mathbf{a} \vert)  \end{equation} where $\vert \mathbf{a} \vert$ is the length of set of image feature vectors extracted from the CNN. This is used to generate the initial hidden and cell states of the LSTM by using an affine transformation followed by a non-linearity ($Tanh$ function) as,
\begin{equation}
    h_0 = Tanh(a_{ave} \star W^h + b^h)
\end{equation}
\begin{equation}
    c_0 = Tanh(a_{ave} \star W^c + b^c)
\end{equation}
where $W^h$, $W^c$ and $b^h$, $b^c$ are weights and biases of the MultiLayer Perceptron (MLP) which is used to model the transformations. 

The successive hidden and cell states are generated during training. Since the generation of words is dependent on the previous words in the sentence as depicted in Equation \ref{eq2}, this dependence can be modelled using the hidden state of the LSTM (which is also modulated by the cell state). Hence,
\begin{equation}\label{eqbig}
    P_\theta (w_i \vert I, w_1, w_2, ..., w_{i-1}) = P_\theta (w_i \vert I, h_i) = f_\theta (w_i, I, h_i) 
\end{equation}
where $f_\theta$ is any differentiable function and since it is recursive in nature it can be modelled using an RNN. Since the hidden state also depends on the previous hidden states, it can be modelled as a function of previous hidden state and inputs as,
\begin{equation}
    h_i = f_\theta (w_{i-1}, h_{i-1}, I)
\end{equation}
where $f_\theta$ is the same differentiable function as in Equation \ref{eqbig} since the model is trained end-to-end with the same parameters. And words are represented as word embeddings which is a function that maps one-hot word vectors to the embedding dimensions and is also learned with the rest of the model, as 
\begin{equation}
    w_{i}^{e} = f_\theta(w_i)
\end{equation}
where $f_\theta$ is the same differentiable function in Equation \ref{eqbig} and $w_{i}^{e}$ is the word embedding vector for word $w_i$.

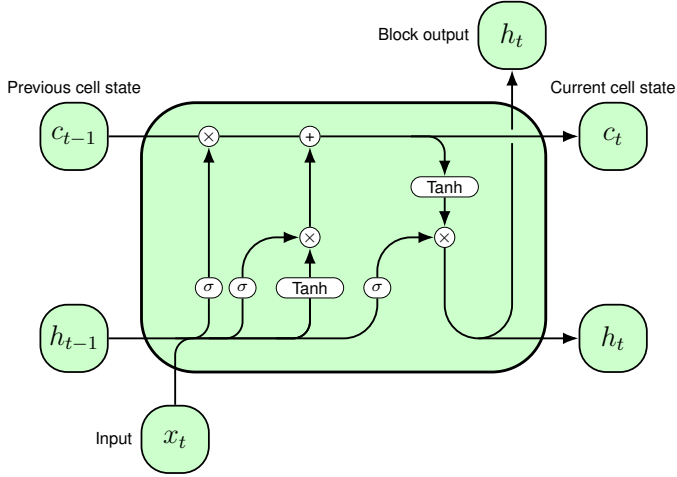
\begin{figure}[h]
\centering
\resizebox{0.5\textwidth}{!}{
\begin{tikzpicture}[
    font=\sf \scriptsize,
    >=LaTeX,
    cell/.style={
        rectangle, 
        rounded corners=8mm,
        draw,
        very thick,
        fill=green!20
        },
    operator/.style={
        circle,
        draw,
        inner sep=-0.5pt,
        minimum height =.3cm,
        fill=white!100,
        },
    function/.style={
        ellipse,
        draw,
        inner sep=1pt
        },
    ct/.style={
        rectangle,
        draw,
        rounded corners=4mm,
        line width = .75pt,
        minimum width=1cm,
        minimum height=1cm,
        inner sep=1pt,
        fill=green!20},
    gt/.style={
        rectangle,
        rounded corners=1.5mm,
        draw,
        minimum width=4mm,
        minimum height=3mm,
        inner sep=1pt,
        fill=white!15},
    mylabel/.style={
        font=\scriptsize\sffamily
        },
    ArrowC1/.style={
        rounded corners=.25cm,
        thick,
        },
    ArrowC2/.style={
        rounded corners=.5cm,
        thick,
        },
    ]

    \node [cell, minimum height =4cm, minimum width=6cm] at (0,0){} ;

    \node [gt] (ibox1) at (-2,-0.75) {$\sigma$};
    \node [gt] (ibox2) at (-1.5,-0.75) {$\sigma$};
    \node [gt, minimum width=1cm] (ibox3) at (-0.5,-0.75) {Tanh};
    \node [gt] (ibox4) at (0.5,-0.75) {$\sigma$};

    \node [operator] (mux1) at (-2,1.5) {$\times$};
    \node [operator] (add1) at (-0.5,1.5) {+};
    \node [operator] (mux2) at (-0.5,0) {$\times$};
    \node [operator] (mux3) at (1.5,0) {$\times$};
    \node [gt, minimum width=1cm] (func1) at (1.5,0.75) {Tanh};

    \node[ct, label={[mylabel]Previous cell state}] (c) at (-4,1.5) {\large{${c}_{t-1}$}};
    \node[ct, label={[mylabel]}] (h) at (-4,-1.5) {\large{${h}_{t-1}$}};
    \node[ct, label={[mylabel]left:Input}] (x) at (-2.5,-3) {\large{${x}_{t}$}};

    \node[ct, label={[mylabel]Current cell state}] (c2) at (4,1.5) {\large{${c}_{t}$}};
    \node[ct, label={[mylabel]}] (h2) at (4,-1.5) {\large{${h}_{t}$}};
    \node[ct, label={[mylabel]left:Block output}] (x2) at (2.5,3) {\large{${h}_{t}$}};

    \draw [ArrowC1] (c) -- (mux1) -- (add1);
    \draw [->,ArrowC2] (add1) -- (c2);
    
    \draw [ArrowC2] (h) -| (ibox4);
    \draw [ArrowC1] (h -| ibox1)++(-0.5,0) -| (ibox1); 
    \draw [ArrowC1] (h -| ibox2)++(-0.5,0) -| (ibox2);
    \draw [ArrowC1] (h -| ibox3)++(-0.5,0) -| (ibox3);
    \draw [ArrowC1] (x) -- (x |- h)-| (ibox3);
    
    \draw [->, ArrowC2] (ibox1) -- (mux1);
    \draw [->, ArrowC2] (ibox2) |- (mux2);
    \draw [->, ArrowC2] (ibox3) -- (mux2);
    \draw [->, ArrowC2] (ibox4) |- (mux3);
    \draw [->, ArrowC2] (mux2) -- (add1);
    \draw [->, ArrowC1] (add1 -| func1)++(-0.5,0) -| (func1);
    \draw [->, ArrowC2] (func1) -- (mux3);

    \draw [->, ArrowC2] (mux3) |- (h2);
    \draw (c2 -| x2) ++(0,-0.1) coordinate (i1);
    \draw [-, ArrowC2] (h2 -| x2)++(-0.5,0) -| (i1);
    \draw [->, ArrowC2] (i1)++(0,0.2) -- (x2);

\end{tikzpicture}}
\caption{Illustration of a basic LSTM cell.}
\label{lstm_diagram}
\end{figure}

We use LSTM as described in \cite{lstm}. The LSTM has three control gates: input, forget and update gates. The equations for updating the different gates are as follows:
\begin{equation}\label{eqinp}
i_t = \sigma (W_ix_t+ R_ih_{t-1}+ b_i)    
\end{equation}
\begin{equation}\label{eqfgt}
f_t = \sigma (W_fx_t+ R_fh_{t-1}+ b_f)    
\end{equation}
\begin{equation}\label{eqopt}
o_t = \sigma (W_ox_t+ R_oh_{t-1}+ b_o)    
\end{equation}
\begin{equation}\label{eqcell}
c_t = f_t \odot c_{t-1} + i_t\odot tanh (W_zx_t+ R_zh_{t-1}+ b_z)     
\end{equation}
\begin{equation}\label{eqhlstm}
h_t = o_t \odot tanh(c_t)   
\end{equation}

where $W_i$ and $R_i$, $W_f$ and $R_f$, $W_o$ and $R_o$ and $W_z$ and $R_z$ are weight matrices (input and recurrent weight matrices) pairs for the input, forget, output and the input modulator(tanh) gates, respectively. $b$ is the bias vector and $\sigma$ is the sigmoid function. It is expressed as $\sigma(x)=1/1+exp(x)$ and condenses the input to the range of (0,1). $\tanh$ is the is hyperbolic tangent function which condenses the input in the range (-1,1). $i_t$, $o_t$ and $f_t$ are input, output and  forget gates respectively. The input gate processes the input information. The output gate generates output based on the input and some of this information has to be dropped which is decided by the cell state. The cell state stores information about the context. The forget gate decides what contextual information has to be dropped from the cell state. The internal structure of the LSTM has been depicted in Figure \ref{lstm_diagram}.

\subsection{CNN + LSTM + Attention method}
\label{cnn_lstm_attention}
In this method, in addition to the the initial time-step, the image information is fed into the LSTM at each time-step. However a separate attention mechanism generates information which is extracted from only certain regions of image which are relevant at the current time-step.

The attention mechanism produces a context vector which represents the relevant portion of the image at each time-step. First a set of weights are calculated for each image feature vector \begin{math}a_i \in \mathbf{a}, i \in (1,2,3,...,\vert a \vert) \end{math} as described in Section \ref{image_feature_extraction}.
\begin{equation}
\begin{split}
    P = \{p_{ti}\}, \hspace{1cm}
    p_{ti} = f_{att}(a_i, h_{t-1})
\end{split}
\end{equation}
where $i \in (1,2,3,...,\vert a \vert)$. Then the attention weights are calculated as,
\begin{equation}
    \boldsymbol{\alpha} = \{ \alpha_{ti} \}, \hspace{1cm} \alpha_{ti} = \frac{exp(p_{ti})}{\sum_{k=1}^n exp(p_{tk})}
\end{equation}
where $\boldsymbol{\alpha}$ is the set of weights, one for each image feature vector $a_i$ in $\mathbf{a}$ such that \begin{math} \sum_{k=1}^{\vert a \vert} \alpha_i = 1 \end{math}.

Then the context vector is calculated by another function, 
\begin{equation}
    \mathbf{z_i} = \Phi(\{ a_i \}, \{ \alpha_i\}) 
\end{equation}
We have used the function $f_{att}$ and $\Phi$ as desrcibed in \cite{attendtell}.

With the context vector thus obtained, the equations for the gates of the LSTM decoder would be,

\begin{equation}\label{eqinp}
i_t = \sigma (W_ix_t+ R_ih_{t-1}+ Z_iz_t + b_i)    
\end{equation}
\begin{equation}\label{eqfgt}
f_t = \sigma (W_fx_t+ R_fh_{t-1}+ Z_fz_t + b_f)    
\end{equation}
\begin{equation}\label{eqopt}
o_t = \sigma (W_ox_t+ R_oh_{t-1}+ Z_oz_t + b_o)    
\end{equation}
\begin{equation}\label{eqcell}
c_t = f_t \odot c_{t-1} + i_t\odot tanh (W_cx_t+ R_ch_{t-1}+ Z_cz_t + b_c)   
\end{equation}
\begin{equation}\label{eqhlstm}
h_t = o_t \odot tanh(c_t)   
\end{equation}
where $W_i$ and $R_i$, $W_f$ and $R_f$, $W_o$ and $R_o$ and $W_c$ and $R_c$ are weight matrices (input and recurrent weight matrices) pairs for the input, forget, output and the input modulator(tanh) gates, respectively. $b$ is the bias vector and $\sigma$ is the sigmoid function.

\section{Experiments and Results}
\label{results}
In this section we describe the experimental details and the results.
We have evaluated Squeezenet \cite{squeezenet}, Shufflenet \cite{shufflenet}, Mobilenet \cite{mobilenet}, MnasNet \cite{mnasnet}, ResNet \cite{resnet}, GoogLeNet \cite{googlenet}, DenseNet \cite{densenet}, Inceptionv4 \cite{inception}, AlexNet \cite{alexnet}, DPN (Dual Path Network) \cite{dpn}, ResNext \cite{resnext}, SeNet \cite{senet}, PolyNet \cite{polynet}, WideResNet \cite{wideresnet}, VGG \cite{vgg16}, NASNetLarge \cite{nasnetamobile} and InceptionResNetv2 \cite{inceptionresnet} CNN models. Out of these we have evaluated five versions of ResNet, viz., Resnet18, ResNet34, ResNet50, Resnet101, Resnet152, four versions of DenseNet, viz., Densenet121, Densenet169, DenseNet201 and Densenet161 and four versions of VGG, viz., VGG-11, VGG-13, VGG-16 and VGG-19 which are similar in architecture but differ widely in terms of number of parameters and also in terms of accuracy and error rates on Object Recognition task with ImageNet dataset. 

We have evaluated the performance using BLEU, METEOR, CIDER, ROUGE-L and SPICE metrics that were recommended in MSCOCO Image caption Evaluation task \cite{coco}. The evaluation results are provided in Tables \ref{cnn_lstm_table} and \ref{cnn_lstm_attention_table} for 'CNN+LSTM' and 'CNN+LSTM+Attention' methods respectively. In addition we have provided some examples of generated captions in Tables \ref{show_tell_examples} and \ref{show_attend_tell_examples} for both the methods.

We have used Flickr8k \cite{flickr8k} dataset which contains around 8000 images with 5 reference captions each. Out of the 8000 images, around 1000 are earmarked for validation set, around 1000 are meant for test set and the remaining are for training set.

\begin{table*}[h!]
    \caption{Performance of CNN+LSTM method using different CNN architectures.}
    \label{cnn_lstm_table}
    \centering
    \begin{tabular}{p{2.5cm}|p{1.5cm}|p{1.25cm}|p{1cm}|p{1cm}|p{1cm}|p{1cm}|p{1cm}|p{1cm}|p{1.2cm}|p{1cm}}
    CNN name & Parameters (in thousands) & Top-5 O.D. error & BLEU-1 & BLEU-2 & BLEU-3 & BLEU-4 & METEOR & CIDER & ROUGE-L & SPICE  \\ \hline
    Squeezenet \cite{squeezenet}&1,248 & 19.58 &60.04 &40.65 & 26.95 & 17.61 & 18.12 & 42.87 & 44.05 & 12.44\\ \hline
    Shufflenet\cite{shufflenet} & 2,279 & 11.68 & 59.70 & 41.18 & 27.84 & 18.67 & 18.24 & 44.36 & 43.66 & 12.61 \\ \hline
    Mobilenet\cite{mobilenet} &3,505 & 9.71 & 60.60 & 41.72 & 28.44 & 18.87 & 18.83 & 47.97 & 44.28 & 13.50 \\ \hline
    MnasNet\cite{mnasnet} &4,383 & 8.456 & 61.19 & 43.02 & 29.43 & 20.10 & 18.94 & 48.19 & 44.88 & 13.46 \\ \hline
    Densenet121  \cite{densenet} &7,979 & 7.83&  61.62 & 43.36 & 29.47 & 19.88 & 19.39 & 48.99 & 45.32 & 13.64 \\ \hline
    ResNet18 \cite{resnet} & 11,689 & 10.92  & 62.21 & 43.45 & 29.84 & 20.30 & 18.91 & 48.31 & 45.33 & 13.49\\ \hline
    GoogLeNet \cite{googlenet} &13,005& 10.47 & 60.69 & 41.57 & 28.20 & 18.91 & 18.66 & 46.42 & 44.38 & 13.01 \\ \hline
    Densenet169  \cite{densenet} &14,150 & 7.00& 63.73 & 45.00 & 30.87 & 21.13 & 19.95 & 52.88 & 46.41 & 14.32\\ \hline
    DenseNet201 \cite{densenet} & 19,447 & 6.43 & 63.29 & 45.11 & 31.36 & 21.63 & 19.80 & 52.21 & 46.40 & 14.16\\ \hline
    Resnet34 \cite{resnet} & 21,798 & 8.58 & 61.08 & 42.69 & 29.32 & 19.98 & 18.98 & 49.78 & 45.01 & 13.32 \\ \hline
    Resnet50 \cite{resnet} & 25,557 & 7.13 & 61.86 & 43.79 & 30.10 & 20.27 & 19.11 & 50.86 & 45.76 & 13.89 \\ \hline
    Densenet161  \cite{densenet} &28,681 & 6.20& 63.12 & 44.68 & 30.76 & 20.79 & 20.00 & 54.24 & 46.19 & 14.26 \\ \hline
    Inceptionv4 \cite{inception} & 42,680 & 4.80 & 59.49 & 40.47 & 27.00 & 18.03 & 18.22 & 43.17 & 43.61 & 12.23\\ \hline
    Resnet101 \cite{resnet} & 44,549 & 6.44 & 62.77 & 44.11 & 30.62 & 21.10 & 19.65 & 53.00 & 45.91 & 14.04 \\ \hline
    InceptionResNetv2 \cite{inceptionresnet} &54,340& 4.9 & 59.50 & 40.55 & 27.36 & 18.21 & 18.79 & 46.35 & 43.54 & 12.90 \\ \hline
    ResNet152 \cite{resnet} & 60,193 & 5.94 & 62.30 & 44.24 & 30.84 & 21.21 & 19.50 & 55.10 & 46.14 & 14.20\\ \hline
    AlexNet \cite{alexnet} & 61,101 & 20.91 & 59.24 & 40.17 & 26.82 & 17.87 & 17.51 & 41.09 & 42.79 & 11.78 \\ \hline
    DPN131 \cite{dpn} & 75,360 & 5.29 & 59.60 & 40.69 & 27.58 & 18.86 & 18.00 & 42.36 & 43.15 & 12.67 \\ \hline
    ResNext101 \cite{resnext} & 88,791 & 5.47 & 62.38 & 43.79 & 29.85 & 20.20 & 19.54 & 51.37 & 45.54 & 14.05 \\ \hline
    NASNetLarge \cite{nasnetamobile} & 88,950 & 3.8 & 56.08 & 36.76 & 23.54 & 15.46 & 16.76 & 34.74 & 40.50 & 11.56\\ \hline
    SeNet154 \cite{senet} &115,089 & 4.47 & 61.67 & 43.18 & 29.72 & 20.19 & 19.48 & 49.89 & 45.24 & 13.95  \\ \hline
    PolyNet \cite{polynet} & 118,733 & 4.25 & 60.26 & 41.26 & 27.68 & 18.68 &  18.02 & 44.23 & 43.61 & 12.37 \\ \hline
    WideResNet101 \cite{wideresnet} & 126,886 & 5.72 & 61.42 & 42.48 & 28.71 & 19.16 & 18.64 & 46.24 & 44.41 & 13.23 \\ \hline 
    VGG-11(bn) \cite{vgg16} & 132,869 & 11.37 & 61.70 & 43.37 & 30.08 & 20.86 & 19.38 & 48.98 & 45.80 & 13.62 \\ \hline
    VGG-13(bn) \cite{vgg16} & 133,054 & 10.75 & 60.79 & 42.42 & 28.91 & 19.70 & 19.06 & 46.57 & 44.84 & 13.39\\ \hline
    VGG-16(bn) \cite{vgg16} & 138,366 & 8.50 & 60.56 & 41.98 & 28.66 & 19.51 & 19.04 & 48.41 & 44.82 & 13.71 \\ \hline
    VGG-19(bn) \cite{vgg16} & 143,678 & 9.12 & 61.40 & 43.09 & 29.49 & 20.02 & 19.15 & 49.42 & 45.43 & 13.61 \\ \hline
    
    \end{tabular}
\end{table*}

\begin{table*}[h!]
    \caption{Performance of CNN+LSTM+Attention method using different CNN architectures.}
    \label{cnn_lstm_attention_table}
    \centering
    \begin{tabular}{p{2.5cm}|p{1.5cm}|p{1.25cm}|p{1cm}|p{1cm}|p{1cm}|p{1cm}|p{1cm}|p{1cm}|p{1.2cm}|p{1cm}} \hline
    CNN name & Parameters (in thousands) & Top-5 O.D. error & BLEU-1 & BLEU-2 & BLEU-3 & BLEU-4 & METEOR & CIDER & ROUGE-L & SPICE  \\ \hline
    Squeezenet \cite{squeezenet}&1,248 & 19.58 & 60.79 & 42.29 & 28.78 & 19.41 & 18.80 & 46.54 & 44.48 & 12.85 \\ \hline
    Shufflenet\cite{shufflenet} & 2,279 & 11.68 & 62.36 & 43.87 & 30.42 & 21.00 & 19.18 & 49.01 & 45.00 & 13.50 \\ \hline
    Mobilenet\cite{mobilenet} &3,505 & 9.71 & 63.69 & 45.33 & 31.72 & 21.89 & 19.63 & 55.36 & 46.28 & 14.25 \\ \hline
    MnasNet\cite{mnasnet} &4,383 & 8.456 & 63.99 & 45.75 & 32.11 & 22.36 & 19.78 & 54.84 & 46.17 & 14.02 \\ \hline
    Densenet121 \cite{densenet} & 7,979 & 7.83 & 64.11 & 45.67 & 31.76 & 22.07 & 20.43 & 55.85 & 46.74 & 14.91 \\ \hline
    ResNet18 \cite{resnet} & 11,689 & 10.92 & 63.26 & 44.87 & 31.07 & 21.24 & 20.08 & 52.44 & 45.84 & 13.75 \\ \hline
    GoogLeNet \cite{googlenet} &13,005& 10.47 & 62.91 & 44.27 & 30.27 & 20.50 & 19.51 & 50.72 & 46.02 & 13.80 \\ \hline
    Densenet169 \cite{densenet} & 14,150 & 7.00 & 64.48 & 46.17 & 32.28 & 22.30 & 20.81 & 56.25 & 46.82 & 14.93 \\ \hline
    DenseNet201 \cite{densenet} & 19,447 & 6.43 & 64.38 & 46.26 & 32.41 & 22.49 & 20.73 & 59.71 & 47.19 & 15.13 \\ \hline
    Resnet34 \cite{resnet} & 21,798 & 8.58 & 63.36 & 45.28 & 31.88 & 22.23 & 19.88 & 55.35 & 46.17 & 14.40 \\ \hline
    Resnet50 \cite{resnet} & 25,557 & 7.13 & 65.32 & 46.92 & 32.81 & 22.58 & 20.87 & 57.12 & 46.95 & 14.90 \\ \hline
    Densenet161 \cite{densenet} & 28,681 & 6.20 &  65.00 & 46.99 & 32.83 & 22.56 & 20.44 & 56.74 & 47.57 & 14.93 \\ \hline
    Inceptionv4 \cite{inception} & 42,680 & 4.80 & 60.17 & 42.24 & 28.71 & 19.35 & 18.76 & 48.00 & 44.33 & 13.26 \\ \hline
    Resnet101 \cite{resnet} & 44,549 & 6.44 & 64.33 & 45.99 & 32.13 & 22.02 & 20.29 & 56.09 & 46.58 & 14.80 \\ \hline
    InceptionResNetv2 \cite{inceptionresnet} &54,340& 4.9 &61.46 & 42.98 & 29.20 & 19.84 & 19.20 & 49.83 & 44.44 & 13.81 \\ \hline
    ResNet152 \cite{resnet} & 60,193 & 5.94 & 65.26 & 47.55 & 33.72 & 23.67 & 20.94 & 58.33 & 47.54 &  15.18 \\ \hline
    AlexNet \cite{alexnet} & 61,101 & 20.91 & 59.93 & 40.97 & 27.80 & 19.06 & 18.67 & 46.11 & 44.09 & 12.57\\ \hline
    DPN131 \cite{dpn} & 75,360 & 5.29 & 62.68 & 44.17 & 30.47 & 20.53 & 19.41 & 49.98 & 45.51 & 13.95 \\ \hline
    ResNext101 \cite{resnext} & 88,791 & 5.47 & 64.78 & 46.07 & 32.36 & 24.45 & 20.93 & 57.67 & 40.04 & 15.28 \\ \hline
    NASNetLarge \cite{nasnetamobile} & 88,950 & 3.8 & 63.60 & 44.66 & 30.16 & 19.93 & 19.73 & 51.34 & 45.49 & 14.00 \\ \hline
    SeNet154 \cite{senet} &115,089 & 4.47 & 64.23 & 45.94 & 32.54 & 22.62 & 20.81 & 58.45 & 46.83 & 15.05 \\ \hline
    PolyNet \cite{polynet} & 118,733 & 4.25 & 62.56 & 44.78 & 31.16 & 21.48 & 19.75 & 53.38 & 45.96 & 13.81 \\ \hline
    WideResNet101 \cite{wideresnet} & 126,886 & 5.72 & 63.47 & 45.37 & 31.71 & 21.73 & 19.84 & 54.27 & 46.23 & 14.51 \\ \hline 
    VGG-11(bn) \cite{vgg16} & 132,869 &11.37 &63.00 & 44.66 & 31.18 & 21.68 & 19.79 & 52.24 & 46.42 & 14.08 \\ \hline
    VGG-13(bn) \cite{vgg16} & 133,054 & 10.75 & 63.64 & 45.09 & 31.26 & 21.41 & 20.25 & 55.17 & 46.35 & 14.64 \\ \hline
    VGG-16(bn) \cite{vgg16} & 138,366 & 8.50 & 63.81 & 45.77 & 32.35 & 22.55 &  20.19 & 55.13 & 46.72 & 14.49 \\ \hline
    VGG-19(bn) \cite{vgg16} & 143,678 & 9.12 &62.57 & 44.63 & 30.97 & 21.44 & 19.76 & 54.10 & 46.23 & 14.44 \\ \hline

    \end{tabular}
\end{table*}

\begin{table*}[h!]
 \caption{Examples of generated captions by CNN+LSTM method using different CNN architectures.}
    \label{show_tell_examples}
    \centering
    \begin{tabular}{p{0.75cm}p{3cm}p{3.0cm}p{3cm}p{3.0cm}p{3cm}} \hline
  Choice of CNN & \includegraphics[height = 2.25cm, width = 3.0 cm]{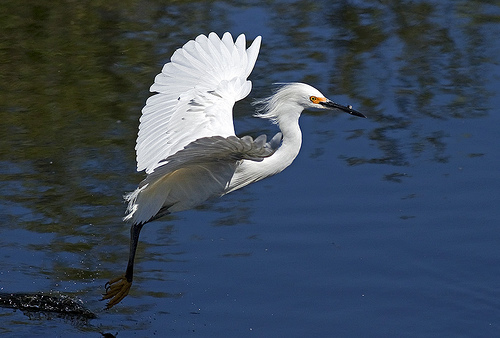}& \includegraphics[height = 2.25cm, width = 3.0 cm]{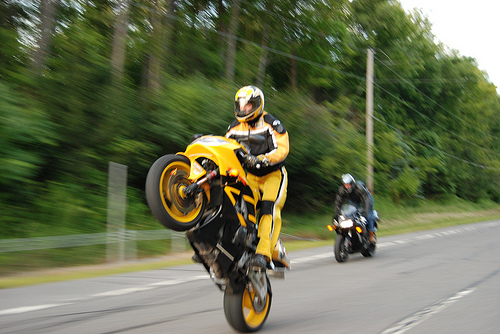} & \includegraphics[height = 3.5cm, width = 3 cm]{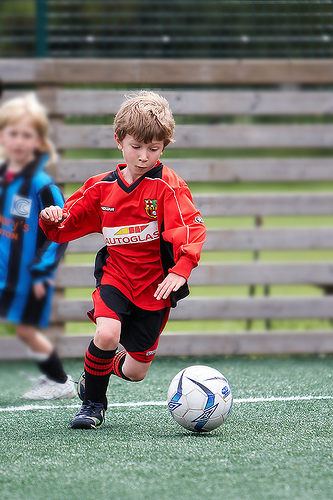} & \includegraphics[height = 2.5cm, width = 3.0 cm]{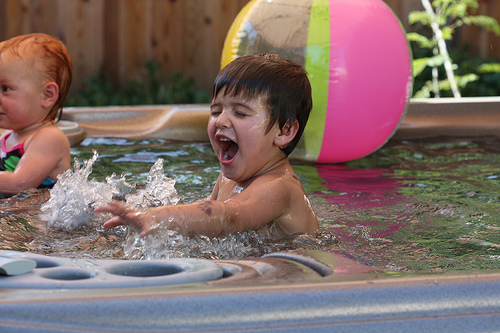} & \includegraphics[height = 3.5cm, width = 3 cm]{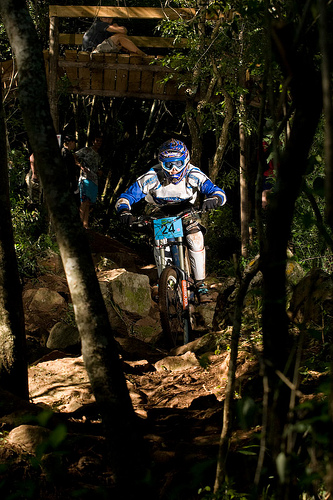} \\ \hline
    ResNet-152  & a white crane flies over the water& a man riding a motorcycle& two young boys playing soccer&two children playing in a pool&a person riding a bike in the woods\\ \hline
Inception-ResNet &a white crane flies over the water&a man riding a motorcycle&two young boys playing soccer&two children are playing in a pool&a man in a blue shirt is riding a bike through a wooded area\\ \hline
NASNET Large &a white crane flies over the water&a man is riding a red motorcycle&a boy in a red uniform kicks a soccer ball&a child plays in a pool&a man on a bike in a forest\\ \hline
VGG-16 &a white bird flies through the water&a man riding a yellow motorcycle&a boy in a soccer uniform kicking a soccer ball&a boy in a blue shirt plays with a plastic toy&a dirt bike rider is airborne in the woods\\ \hline
Alexnet &a white bird flies over the water&a man in yellow and yellow motorcycle&a boy in a red uniform runs with a soccer ball&a young girl in a bathing suit is jumping into a pool&a man is riding a bike on a dirt path\\ \hline
Squeezenet &a white bird in the water&a man in a yellow helmet is riding a bike&a boy in a red and white uniform is playing soccer&a little girl in a pink dress is playing in a pool&a person riding a bike through the woods\\ \hline
Densenet-201 &a white bird flies over the water&two bikers racing on the road&two children playing soccer&a young boy in a pool&a man on a bike is riding a bike through the woods\\ \hline
GoogLeNet & \hspace{0.05cm} a white bird flies through the water&a man on a motorcycle is riding on a street&a young boy wearing a red shirt and a blue soccer ball&a little boy is being splashed in a pool&a man is riding a bike through the woods\\ \hline
Shufflenet &a white bird flies through the water&a man in a yellow helmet riding a yellow bike&a little boy in a red shirt is playing with a soccer ball&two young children playing in a fountain&a man in a blue helmet rides a bike through the woods\\ \hline
Mobilenet &a white bird is flying over water&a person riding a bike in a race&a boy in a red and white uniform is playing soccer&a young boy in a swimming pool&a person riding a dirt bike in the woods\\ \hline
Resnext-101 &a white bird flies over the water&a man on a motorcycle is riding a motorcycle&a soccer player in a red uniform kicks a soccer ball&a little girl is playing in a pool&a dirt bike rider in the woods\\ \hline
Wide ResNet-101 &a white bird flies over the water&a man riding a motorcycle&two boys playing soccer on a field&a boy is splashing in a pool&a person riding a dirt bike through the woods\\ \hline
Mnasnet &a white bird in the water&a man in a yellow jacket rides a motorcycle&a boy in a blue uniform is playing soccer&a little boy is playing in a pool&a man on a bike in the woods\\ \hline
Inception&a white bird flying over water&a man is riding a bike on a track&two boys playing soccer&two children play in a pool&a person in a blue shirt and blue jeans is sitting on a tree\\ \hline
DPN-131&a white crane landing in the water&a person on a motorcycle&a young boy in a soccer uniform kicking a soccer ball&a little boy in a swimming pool&a person is riding a bike in the woods\\ \hline
Senet-154&a white crane flying over water&a man is riding a yellow motorcycle&a man in a red uniform kicking a soccer ball&a little boy in a swimming pool&a person rides a bike through the woods\\ \hline
Polynet&a white bird flies over the water&a man rides a motorcycle&a boy in a blue uniform is chasing a soccer ball&a girl in a pink shirt is playing in a kiddie pool&a person rides a bike through the woods\\ \hline
    \end{tabular}
\end{table*}

\begin{table*}[h!]
    \caption{Examples of generated captions by CNN+LSTM+Attention method using different CNN architectures.}
    \label{show_attend_tell_examples}
    \centering
    \begin{tabular}{p{0.75cm}p{3cm}p{3.0cm}p{3cm}p{3.0cm}p{3cm}} \hline
      Choice of CNN & \includegraphics[height = 2.25cm, width = 3.0 cm]{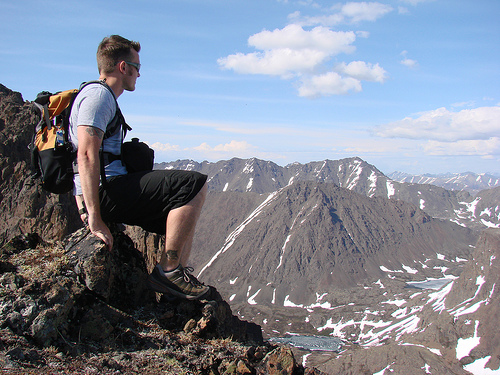} & \includegraphics[height = 3.5cm, width = 3 cm]{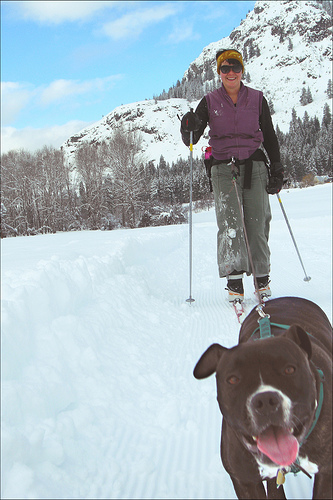} & \includegraphics[height = 2.25cm, width = 3.0 cm]{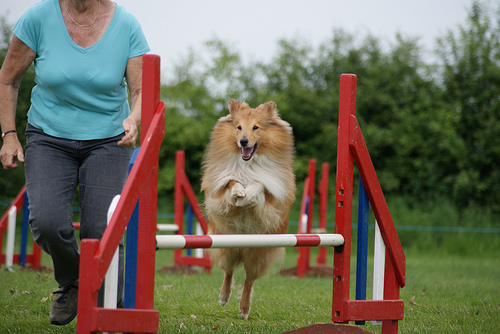} & \includegraphics[height = 2.25cm, width = 3.0 cm]{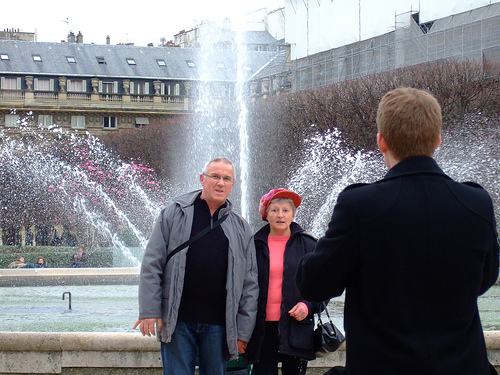} & \includegraphics[height = 3.5cm, width = 3 cm]{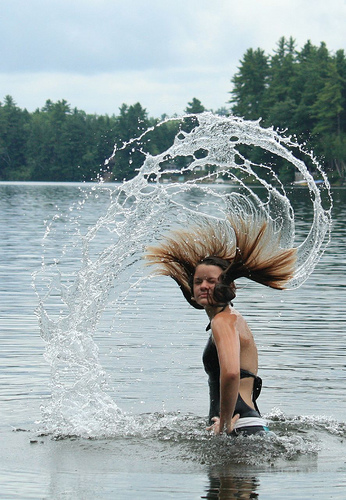}\\ \hline
       ResNet-152  & a man is standing in front of a mountain & a dog runs through the snow & a dog jumps over a hurdle &a man and a woman are sitting on a fountain& a young girl in a pink bathing suit is playing in the water\\ \hline
       Inception-ResNet  & a man with a backpack stands on a mountaintop &a man in a red jacket is skiing down a snowy hill & a brown dog jumps over a hurdle &two children playing in a fountain& a little girl plays in the water\\ \hline
       NASNET Large & a man sits on top of a mountain &a dog is running through the snow & a brown dog is jumping over a hurdle &a group of people are playing in a fountain& a girl in a pink swimsuit is jumping into the water \\ \hline
       VGG-16 & a man is standing on top of a mountaintop & a brown dog is standing in the snow & a dog is jumping over a hurdle &a group of people are sitting on a ledge overlooking a city& a woman in a swimsuit is standing in the water\\ \hline
       Alexnet & a man is standing on top of a mountain &a brown dog is running through snow&a dog jumps over a hurdle &a man in a black jacket is standing next to a building& a boy in a pool \\ \hline
       Squeezenet & a group of people sit on a snowy mountain & a man in a red jacket is standing on a snowy hill& a brown and white dog with a red and white dog &a group of people stand in front of a building& a woman in a white shirt is walking through the water \\ \hline
       Densenet-201 & a man in a blue shirt is standing in the mountains&a brown dog is jumping in the snow& a dog jumps over a hurdle &a man and a woman are standing in front of a fountain& a young girl jumping into the water \\ \hline
       GoogLeNet & \hspace{0.05cm} a man is standing on a mountaintop&a black and white dog is running through the snow& a dog jumps over a hurdle &a group of people stand in a fountain&a young boy plays in the water \\ \hline
       Shufflenet & a man and a woman are sitting on a rock overlooking the mountains&a man in a red jacket is standing on a snowy hill& a woman and a dog are playing in a yard &a man and a woman are walking down a city street&a man is standing on the shore of a body of water\\ \hline
       Mobilenet & a man stands on a mountain&a man is skiing down a snowy hill& a woman and a woman sitting on a bench &two men are standing next to a fountain&a girl in the water\\ \hline
       Resnext-101 & a man with a backpack stands on a mountaintop&a person is skiing down a snowy hill& a dog jumping over a hurdle &a man and a woman are standing in a fountain&a woman in a bikini is playing in the water \\ \hline
       Wide ResNet-101 & a man is standing on top of a mountain&a dog is running through the snow& a man and a dog on a leash &a group of people are standing in a fountain&a woman in a bathing suit walks along the water\\ \hline
       Mnasnet & a man and a woman are standing in the mountains&a brown dog is running through the snow& a dog jumps over a hurdle &a group of people are standing in front of a fountain&a boy is splashing in the water\\ \hline
       Inception & a man stands on a rock overlooking the mountains&a black and white dog in the snow& a brown and white dog is jumping over a hurdle &a group of people are playing in a fountain&a dog walks through the water\\ \hline
       DPN-131 & a man is standing on top of a mountain&a man and a dog play in the snow& a dog jumps over a hurdle &a group of people stand in a fountain&a girl in a swimsuit is jumping into the water\\ \hline
       Senet-154 & a man is standing in front of a mountain&a dog is running through the snow& a dog jumps over a hurdle &a man is standing in front of a fountain&a girl in a red bathing suit splashes in the water\\ \hline
       Polynet & a man stands on a mountaintop&a dog is jumping over a snowy hill& a dog is jumping over a hurdle &a group of people are standing in front of a fountain fountain&a woman in a bathing suit is standing in front of a waterfall\\ \hline
       
    \end{tabular}

\end{table*}

We can make following observations from the results:
\begin{itemize}
    \item For example, there is a variation of around 4 to 5 points in the evaluation metrics between the best and worst performing models in both Tables \ref{cnn_lstm_table} and \ref{cnn_lstm_attention_table}.
    \item In addition, the performance of a decoder framework which employs additional methods of guidance (such as attention) but uses a lower performing encoder can be worse than simpler methods which use better performing CNN encoder. For example, the best performing model using CNN+LSTM method (Table \ref{cnn_lstm_table}) have better performance than lower performing models using CNN+LSTM+Attention method (Table \ref{cnn_lstm_attention_table}).
    \item Although different variants of the same model (such as ResNet, Densenet and VGG) differ greatly with respect to the number of parameters, they generate image captioning performances which differ only by around 1 point on most evaluation metrics. ResNet18, being the smallest model in terms of number of parameters (among ResNet based CNNs) performs competitively as compared to the larger ResNet variants which have many times more parameters. We also observe that DenseNet121 and VGG-11 being the smallest models among DenseNet  and VGG models, respectively, outperform other DenseNet and VGG based CNNs in evaluation scores along certain metrics. 
    \item Also the different variants of ResNet \cite{resnet}, VGG \cite{vgg16} and DenseNet \cite{densenet} architectures differ greatly in terms of Top-5 error on Object Detection task when evaluated with Imagenet dataset. However, that difference does not translate to similar difference in performance in Image captioning task.   
    \item For each image, most models generate reasonable captions but there is a great variation in the caption sentences generated with different models. In some cases, captions generated with different models describe different portions of the image and sometimes some models focus on a certain object in the image instead of providing a general overview of the scene.
    \item In some cases, models do not recognize certain objects in the image. In particular, we have observed many cases of incorrect gender identification which points out to possible statistical bias in the dataset towards a particular gender in a certain context.
\end{itemize}
Thus we can conclude that choice of CNN for the encoder significantly influences the performance of the model. In addition to the general observations, we are able to deduce the following specific observations about the choice of CNN:
\begin{itemize}
    \item ResNet\cite{resnet} and DenseNet\cite{densenet} CNN architectures are well suited to Image caption generation and generate better results while having a lower model complexity than other architectures.
\end{itemize}

\section{Conclusion}
\label{conclusion}
In this work, we have evaluated encoder-decoder and attention based caption generation frameworks with different choices of CNN encoders and observed that there is a wide variation in terms of both the scores, as evaluated with commonly used metrics (BLEU, METEOR, CIDER, SPICE, ROUGE-L), and also the generated captions while using different CNN encoders. In terms of most metrics, there is a difference in performance of around 4-5 points between the worst and best performing models. Hence, the choice of particular CNN architecture plays a big role in the image caption generation process. In particular, ResNet and DenseNet based CNN architectures lead to better overall performance while at the same using lesser parameters than other models.

Also, since there is a great variation in the generated captions for each image, it may be possible to use ensemble of models, each of which utilize a different CNN as encoder, to increase diversity of generated captions. Also, model ensembling would lead to better performance. In the works proposed in the literature, model ensembling has been used such as in \cite{vinyals} but such model ensembles utilize similar models trained with different hyperparameters. Using ensembles of models, which use different CNN encoders is an area which could be explored in future works.

Furthermore, we hope that this analysis of the effect of choice of different CNNs for image captioning will aid the researchers in better selection of CNN architectures to be used as encoders in image feature extraction for Image Caption Generation.

\section*{Acknowledgment}
We are greatly indebted to the MultiMedia Processing and Language Processing Laboratories at the Department of Computer Science and Engineering, National Institute of Technology, Silchar, India for providing us the GPU-equipped workstations which were indispensable for this work. Also, the Office of Head of Department, Department of Computer Science and Engineering at National Institute of Technology, Silchar also provided one GPU equipped workstation for this work for which we are greatly obliged.

This work was not supported by any financial grant and there do not exist any conflicts of interest.

\end{document}